\documentclass[conference]{IEEEtran}
\IEEEoverridecommandlockouts
\usepackage{cite}
\usepackage{amsmath,amssymb,amsfonts}
\usepackage{algorithm}
\usepackage{algorithmic}
\usepackage{graphicx}
\usepackage{textcomp}
\usepackage{xcolor}
\usepackage{subcaption}
\usepackage{amsthm}
\usepackage{graphicx}
\usepackage{multirow}
\usepackage{subcaption}
\usepackage{setspace}
\usepackage{booktabs}

\def\BibTeX{{\rm B\kern-.05em{\sc i\kern-.025em b}\kern-.08em
    T\kern-.1667em\lower.7ex\hbox{E}\kern-.125emX}}

\begin{document}

\theoremstyle{definition}
\newtheorem{definition}{Definition}

\newcommand{\algname}{MSGAT}
\newcommand{\Poincare}{Poincar\'e }
\newcommand{\Mobius}{M\"obius }
\newcommand{\jong}[1]{\textcolor{blue}{#1}}

\author{\IEEEauthorblockN{Jongmin Park, Seunghoon Han, Jong-Ryul Lee\IEEEauthorrefmark{1}\thanks{\IEEEauthorrefmark{1}Corresponding authors.}, Sungsu Lim\IEEEauthorrefmark{1}}
\IEEEauthorblockA{Department of Computer Science and Engineering, Chungnam National University, Daejeon, South Korea} 
\{pa5398, tmdgns129\}@g.cnu.ac.kr, \{jongryul.lee, sungsu\}@cnu.ac.kr}

\title{Multi-Hyperbolic Space-based Heterogeneous Graph Attention Network}

\maketitle

\begin{abstract}
To leverage the complex structures within heterogeneous graphs, recent studies on heterogeneous graph embedding use a hyperbolic space, characterized by a constant negative curvature and exponentially increasing space, which aligns with the structural properties of heterogeneous graphs. However, despite heterogeneous graphs inherently possessing diverse power-law structures, most hyperbolic heterogeneous graph embedding models use a single hyperbolic space for the entire heterogeneous graph, which may not effectively capture the diverse power-law structures within the heterogeneous graph. To address this limitation, we propose \textit{Multi-hyperbolic Space-based heterogeneous Graph Attention Network (\algname)}, which uses multiple hyperbolic spaces to effectively capture diverse power-law structures within heterogeneous graphs. We conduct comprehensive experiments to evaluate the effectiveness of \algname{}. The experimental results demonstrate that \algname{} outperforms state-of-the-art baselines in various graph machine learning tasks, effectively capturing the complex structures of heterogeneous graphs.
\end{abstract}


\section{Introduction}
Recently, the demand for effective methods to learn semantic information and complex structures in heterogeneous graphs, which consist of various node and link types, has been steadily increasing due to their ability to represent real-world scenarios. In heterogeneous graphs, metapaths are defined as sequences of node/link types. Leveraging metapaths enables us to capture semantic information and complex structures within such graphs effectively. Accordingly, recent studies~\cite{wang2019heterogeneous, yun2019graph, li2021graphmse} have focused on efficiently learning heterogeneous graph representations by leveraging metapaths.

\begin{figure}[t!]
\centering
    \includegraphics[width=0.7\columnwidth]{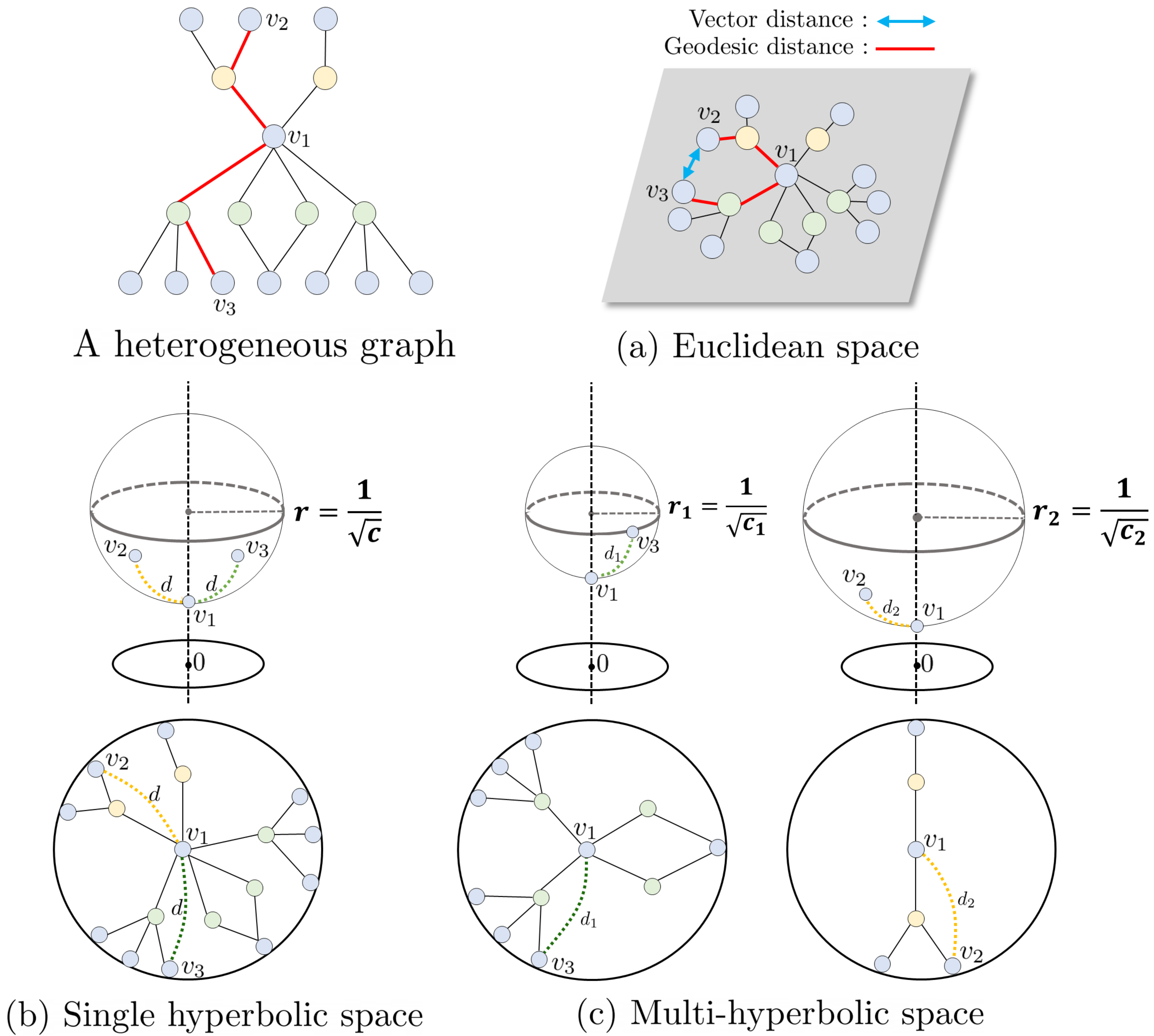}
    \caption{Examples of heterogeneous graph representations in various embedding spaces.}
    \vspace{-2em}
    \label{figure:toy_example}
\end{figure}

Despite their notable achievements, they may struggle to effectively capture complex structures (e.g., power-law structure) within heterogeneous graphs because they use Euclidean space as the embedding space. In heterogeneous graphs, we often observe hierarchical or power-law structures where the number of nodes grows exponentially, corresponding to specific metapaths. Using Euclidean space as the embedding space to learn such complex structures may result in distortions and limitations~\cite{pei2020curvature}. For example, as shown in Figure~\ref{figure:toy_example}(a), given a graph with a hierarchical structure, distortions can occur where the geodesic distance between two nodes ($v_2$ and $v_3$) is far, but the vector distance in the embedding space is represented as close.

Some recent heterogeneous graph embedding models address this challenge by using the hyperbolic space as the embedding space. Compared to Euclidean space, hyperbolic space has a constant negative curvature and grows exponentially. Some recent studies~\cite{nickel2017poincare, balazevic2019multi, pan2021hyperbolic, wang2019hyperbolic, li2023multi, park2024hyperbolic} argue that these inherent properties of hyperbolic space offer a solution to represent complex structures effectively. 
While these studies achieved significant performance, would representing a heterogeneous graph with different complex  structures based on semantic information in a single hyperbolic space be effective? In hyperbolic space, the extent to which hyperbolic space grows exponentially is determined by the negative curvature. Conversely, we can interpret negative curvature as indicating the degree of power-law distribution in hyperbolic space. Therefore, using multiple hyperbolic spaces with distinct negative curvatures that effectively represent each power-law structure within a heterogeneous graph would be more effective for heterogeneous graph representation learning.

\begin{figure}[t!]
\captionsetup[subfigure]{justification=centering}
\centering
        \begin{minipage}[b]{0.45\columnwidth}
                \centering
                \includegraphics[width=\linewidth]{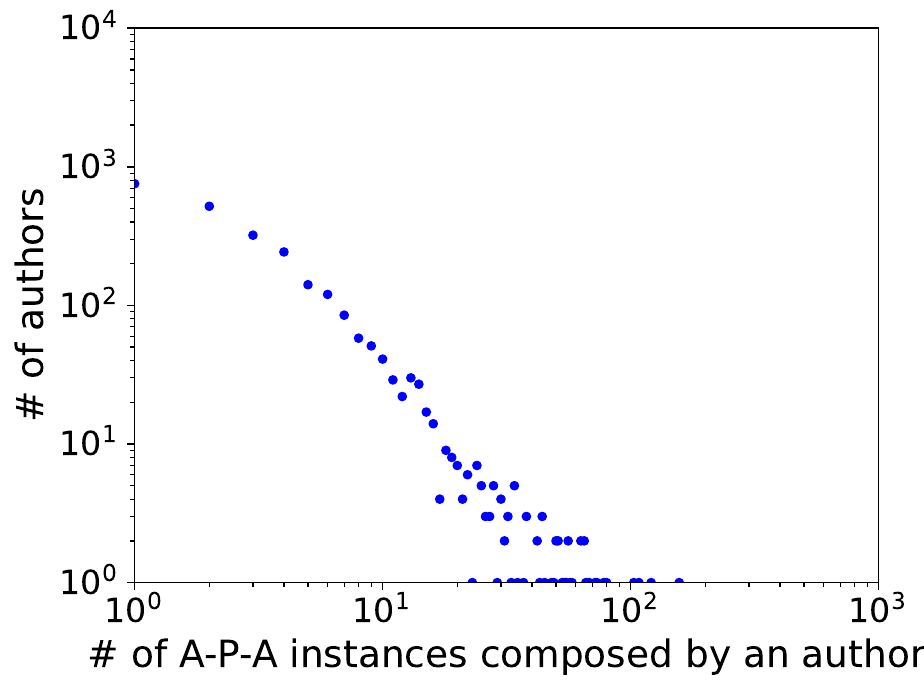}
                \subcaption{A-P-A in DBLP \\($\delta_{avg}$ : 0.7892)}
        \end{minipage}%
        \begin{minipage}[b]{0.45\columnwidth}
                \centering
                \includegraphics[width=\linewidth]{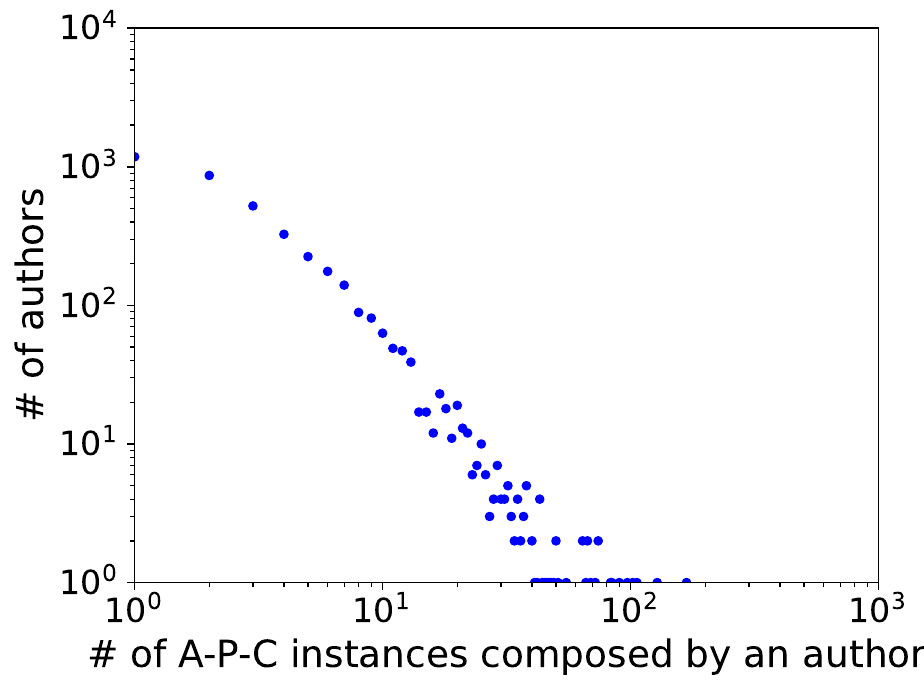}
                \subcaption{A-P-C in DBLP\\($\delta_{avg}$ : 0.3124)}
        \end{minipage}%
        \caption{Metapath instance distributions of some metapaths in DBLP. $\delta$ represents Gromov's $\delta$-hyperbolicity of each metapath-based subgraph.}
        \vspace{-1.75em}
        \label{fig:distributions}
\end{figure}

As illustrated in Figure~\ref{figure:toy_example}(b), if power-law structures with different degree distributions are learned in the same hyperbolic space, it will not be effective in capturing their structural properties. This is because, according to two different metapaths representing different degree distributions, they are represented at the same hyperbolic distance $d$ from the central embedding target node $v_1$ to its metapath-based neighbors. In contrast, as shown in Figure~\ref{figure:toy_example}(c), if structures with more pronounced power-law distributions are learned in hyperbolic space with steeper curvature, and those with less pronounced power-law structures are learned in hyperbolic space with relatively softer curvature, the structural properties of each metapath can be effectively captured. This is because, according to two metapaths representing different degree distributions, they are represented at different distances $d_1$ and $d_2$ from the target node embedding to its metapath-based neighbors.

Additionally, as shown in Figure~\ref{fig:distributions}, in real-world heterogeneous graphs, we can observe multiple power-law structures corresponding to specific metapaths that are similar but distinct. Also, we calculate the average Gromov's $\delta$-hyperbolicity~\cite{adcock2013tree} of each metapath-based subgraph, where lower $\delta$ values indicate that the structure of the subgraph tends to exhibit more hierarchical structures. 
The average Gromov $\delta$-hyperbolicity allows for identifying distinct hierarchical structures in power-law structures with similar distributions, despite their apparent similarity.


Based on these observations, we propose Multi-hyperbolic Space-based heterogeneous Graph Attention Network (\algname{}) to effectively learn various semantic structural properties in heterogeneous graphs with power-law structures. Specifically, \algname{} addresses metapath predefinition challenges by sampling metapath instances and utilizing intra-hyperbolic space attention to learn representations in metapath-specific hyperbolic spaces, where curvatures are learnable parameters. Also, we utilize inter-hyperbolic space attention to aggregate semantic information across distinct metapaths. Through these approaches, our model can effectively learn power-law structures and semantic information within the heterogeneous graph.


The main contributions of our work can be summarized as follows:
\begin{itemize}
    \item We propose a novel hyperbolic heterogeneous graph attention network that uses multiple hyperbolic spaces as embedding spaces for distinct metapaths.
    \item We design graph attention mechanisms in multiple hyperbolic spaces to enhance heterogeneous graph representations, capturing diverse degree distributions and semantic heterogeneity.
    \item The experimental results demonstrate that \algname{} outperforms state-of-the-art baselines in various downstream tasks with heterogeneous graphs.
\end{itemize}

\begin{figure*}[h!]
\centering
    \includegraphics[width=0.7\textwidth]{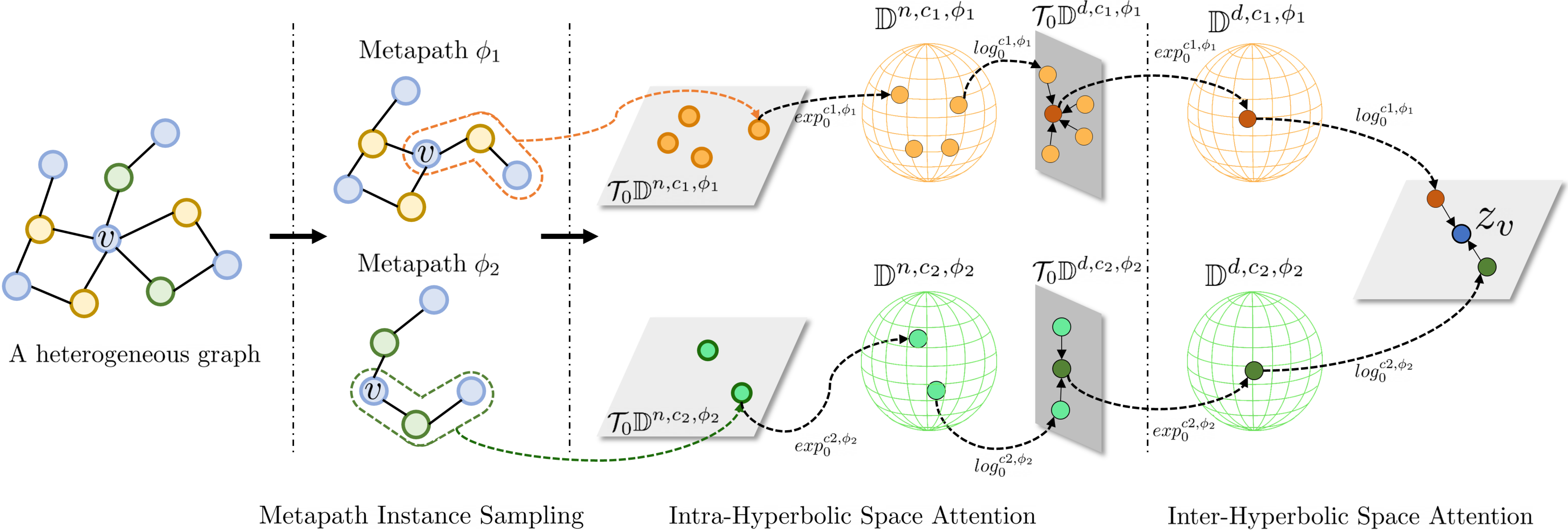}
    \caption{The framework of proposed \algname{}.}
    \vspace{-1.5em}
    \label{figure:framework}
\end{figure*}

\section{Preliminaries}
\subsection{Hyperbolic Space}
\begin{definition}[\bf \Poincare ball model]
\Poincare ball model with curvature $-c\ (c>0)$ is defined by the Riemannian manifold $(\mathbb{D}^{n,c}, g_x^c)$, where
\vspace{-0.5em}
\begin{align*}
    & \mathbb{D}^{n,c}=\{x \in \mathbb{R}^n : c||x||^2 < 1\},\\
    & g_x^c = \left(\lambda^c_x\right)I_d.
\end{align*}
Here, $\mathbb{D}^{n,c}$ is the open $n$-dimensional unit ball with radius $\frac{1}{\sqrt{c}}$ and $g_x^c$ is the Riemannian metric tensor where $\lambda_x^c=\frac{2}{1-c||x||^2}$ and $I_d$ is the identity matrix. We denote $\mathcal{T}_x\mathbb{D}^{n,c}$ as the tangent space centered at point $x$
\end{definition}

\begin{definition}[\bf \Mobius addition] Given the point $x,y \in \mathbb{D}^{n,c}$, \Mobius addition which represents the equation for the addition operation in the \Poincare ball model with curvature $-c\;(c>0)$ is defined as follows:
\begin{align*}
    x\oplus_c y = \frac{\left(1+2c\langle x,y\rangle+c||y||^2\right)x+\left(1-c||x||^2\right)y}{1+2c\langle x,y \rangle+c^2||x||^2||y||^2},
\end{align*}
where $\langle\cdot\rangle$ is the Euclidean inner product and $||\cdot||$ is the Euclidean norm.
\end{definition}

\begin{definition}[\bf Exponential and logarithmic maps]
In \Poincare ball model with curvature $-c\;\left(c>0\right)$, the exponential map $exp^c_x:\mathcal{T}_x\mathbb{D}^{n,c}\rightarrow\mathbb{D}^{n,c}$ and logarithmic map $log^c_x:\mathbb{D}^{n,c}\rightarrow \mathcal{T}_x\mathbb{D}^{n,c}$ are defined as shown below:
\begin{align*}
    & exp_x^c\left(v\right) = x\oplus_c\left(\text{tanh}\left(\sqrt{c}\frac{\lambda_x^c||v||}{2}\right)\frac{v}{\sqrt{c}||v||}\right),\\
    &log_x^c\left(y\right)=\frac{2}{\sqrt{c}\lambda_x^c}\text{tanh}^{-1}\left(\sqrt{c}||-x\oplus_cv||\right)\frac{-x\oplus_c y}{||-x\oplus_c y||},
\end{align*}
\end{definition}
where x and y are the points in 
the hyperbolic space $\mathbb{D}^{n,c}$ and $x\neq y$. $v$ is a nonzero tangent vector in the tangent space $\mathcal{T}_x\mathbb{D}^{n,c}$.

\begin{definition}[\bf Hyperbolic matrix-vector multiplication] 
Given a point $x\in\mathbb{D}^{n,c}$ and a matrix $M\in\mathbb{R}^{m\times n}$, the matrix multiplication operation in hyperbolic space is defined as follows:
\vspace{-1em}
\begin{align*}
    & M\otimes_c x = exp_0^c\left(Mlog_0^c\left(x\right)\right),
\end{align*}
where $\mathbf{0}\in\mathbb{R}^{n}$ is a zero vector. 
\end{definition}

\begin{definition}[\bf Hyperbolic non-linear activation function]
Given the point $x\in\mathbb{D}^{n,c}$, the hyperbolic non-linear activation function is defined as follows:
\vspace{-0.5em}
\begin{align*}
    & \sigma\otimes_c\left(x\right) = exp_0^c\left(\sigma\left(log_0^c\left(x\right)\right)\right),
\end{align*}
where $\sigma$ is the Euclidean non-linear activation function.
\end{definition}

\section{Methodology}
\subsection{Metapath Instance Sampling}
To capture the structural properties within a heterogeneous graph $\mathcal{G}$, we sample a metapath instance set $\mathcal{P}_v$ for a given embedding target node $v$ in $\mathcal{V}_t$. Each metapath instance $p$ in $\mathcal{P}_v$ starts from node $v$ and has a length within a maximum metapath length $l$. We use breadth-first search for this procedure.

\subsection{Intra-Hyperbolic Space Attention}
\subsubsection{Hyperbolic mean-linear encoder}
\label{sec:encoder}
After metapath instance sampling, we compose the hyperbolic mean-linear encoder to transform all the node features within  a metapath instance into a single feature. This transformation function can be formulated as follows:
\vspace{-0.5em}
\begin{align}
     x_p^{\mathbb{E}} &= \frac{1}{j}\sum_{i=1}^j x_i \left(j\leq l\right),\\
     x_p^{\mathbb{H,\phi}} &= W_t\otimes_c exp_0^{c,\phi}\left(x_p^{\mathbb{E}}\right).
     \vspace{-0.25em}
\end{align}
In Equation (1), $x_i \in \mathbb{R}^n$ denotes the features of node $i$, $j$ denotes the length of the metapath instance $p$, $W_t\in\mathbb{R}^{n\times n}$ denotes a transformation matrix, and $x_p^\mathbb{E} \in \mathbb{R}^{n}$ denotes the Euclidean feature of the metapath instance $p$.

In Equation (2), given Euclidean metapath instance feature $x_p^{\mathbb{E}}$, we first map $x_p^{\mathbb{E}}$ to a metapath-specific hyperbolic space $\mathbb{D}^{n,c,\phi}$ via the exponential map $exp_0^{c,\phi}:\mathcal{T}_0\mathbb{D}^{n,c,\phi}\rightarrow\mathbb{D}^{n,c,\phi}$. To adopt exponential map, we assume $x_p^{\mathbb{E}}$ is included in the tangent space $\mathcal{T}_0\mathbb{D}^{n,c,\phi}$ at point $x=0$.
Note that, $ x_p^{\mathbb{H,\phi}}\in\mathbb{D}^{n,c,\phi}$ denotes hyperbolic metapath instance feature. Here, $\mathbb{D}^{n,c,\phi}$ is a metapath-specific hyperbolic space that effectively represents structural properties of metapath instances following a specific metapath $\phi$. 

Additionally, $-c\;\left(c>0\right)$ is a learnable parameter that represents the negative curvature of hyperbolic space, and each metapath-specific hyperbolic space for metapath $\phi$ has a distinct negative curvature.

\subsubsection{Hyperbolic metapath instance embedding} We utilize hyperbolic linear transformation with a hyperbolic non-linear activation function to obtain metapath instance embedding in hyperbolic space. The formulation of this process is as follows:
\vspace{-1em}
\begin{align}
\label{eq:3}
    h_p^{\mathbb{H},\phi} &= \sigma\otimes_c\left(W_1\otimes_c x_p^{\mathbb{H,\phi}}\right)\oplus_c exp_0^{c,\phi}\left(b_1\right).
    \vspace{-1em}
\end{align}

In Equation~(\ref{eq:3}), $h_p^{\mathbb{H},\phi}\in\mathbb{D}^{d,c,\phi}$ is a latent representation of metapath instance $p$ in metapath-specific hyperbolic space $\mathbb{D}^{d,c,\phi}$. Here, $d$ is the dimension of hyperbolic space for latent metapath instance representations. Additionally, $W_1\in\mathbb{R}^{d\times n}$ is a weight matrix and $b_1\in\mathbb{R}^d$ is a bias vector.

\subsubsection{Intra-metapath specific hyperbolic space attention} To aggregate different latent metapath instance representations, we define attention mechanisms in metapath-specific hyperbolic space. First, we calculate the importance of each metapath instance $\alpha_p$ as follows:
\vspace{-0.5em}
\begin{align}
\label{eq:4}
    e_p &= a^T\cdot log_0^{c,\phi}\left(h_p^{\mathbb{H},\phi}\right), \\
    \alpha_p &= \frac{\text{exp}\left(e_p\right)}{\sum_{q\in\mathcal{P}_v^{\phi}}\text{exp}\left(e_q\right)}.
\end{align}

In the above equations, $e_p$ denotes the importance of each metapath instance $p \in \mathcal{P}_v^{\phi}$, where $log_0^{c,\phi}:\mathbb{D}^{d,c,\phi}\rightarrow\mathcal{T}_0\mathbb{D}^{d,c,\phi}$ denotes logarithmic map function and $\mathcal{P}_v^{\phi}$ denotes a subset of $\mathcal{P}_v$ consisting of metapath instances that follow a specific metapath $\phi$ and $a\in\mathbb{R}^{d}$ is an attention vector for metapath instance. After calculating the importance of each metapath instance, we apply the softmax function to these values to obtain the weight of each metapath instance.

Then the metapath-specific embedding for node $v$ is obtained from the weight of each metapath instance and their latent representations in metapath-specific hyperbolic space. This process can be formulated as below:
\vspace{-0.5em}
\begin{align}
\label{eq:6}
    h_v^\phi &= \sigma \otimes_c \bigg( exp_0^{c,\phi} \bigg( \sum_{p \in \mathcal{P}_v^{\phi}} \alpha_p \cdot log_0^{c,\phi} \big( h_p^{\mathbb{H}, \phi} \big) \bigg) \bigg),
\end{align}
where $h_v^\phi\in\mathbb{D}^{d,c,\phi}$ denotes metapath specific embedding for node $v$. Note that, in Equation~(\ref{eq:3}) and~(\ref{eq:6}), $\sigma \otimes_c$ denotes the hyperbolic non-linear activation function with LeakyReLU.

\subsubsection{Hyperbolic multi-head attention} As shown in the equation below, we adopt multi-head attention in hyperbolic space to enhance metapath-specific embeddings and stabilize the learning process. Specifically, we divide the attention mechanisms into $K$ independent attention mechanisms, conduct them in parallel, and then concatenate the metapath-specific embedding from each attention mechanism to obtain the final metapath-specific embedding $h_v^\phi$.
\begin{align}
    h_v^\phi = \parallel_{k=1}^{K} \sigma \otimes_c \bigg( exp_0^{c,\phi} \bigg( \sum_{p \in \mathcal{P}_v^{\phi}} \alpha_p^k \cdot log_0^{c,\phi} \big(h_{p}^{\mathbb{H}, \phi} \big) \bigg) \bigg).
\end{align}

\subsection{Inter-Hyperbolic Space Attention}
\subsubsection{Node embedding space mapping}
Once the metapath-specific embedding $h_v^\phi$ is obtained for each metapath $\phi$ in $\Phi$, we aggregate them using attention mechanisms to learn the importance of each metapath-specific embedding, which represents semantic structural information.

First, we map metapath-specific embeddings to the same embedding space because they are included in different metapath-specific hyperbolic spaces $\mathbb{D}^{d,c,\phi}$ corresponding to specific metapaths $\phi$. Since the curvatures of these metapath-specific hyperbolic spaces are different, it is difficult to aggregate them directly in their respective hyperbolic spaces. Instead, we map metapath-specific embeddings into the same tangent space from different hyperbolic spaces. The mapping operation can be formulated as follows:
\begin{align}
\label{eq:8}
    g_v^{\phi} &= W_{2}\cdot log_0^{c,\phi}\left(h_v^{\phi}\right).
\vspace{-0.5em}
\end{align}

In Equation~(\ref{eq:8}), with the logarithmic map $log_0^{c,\phi}$, we can map $h_v^{\phi}\in\mathbb{D}^{d,c,\phi}$ to tangent space $\mathcal{T}_0\mathbb{D}^{d,c,\phi}$ which is a plane like Euclidean space. Then, with the transformation matrix $W_2\in\mathbb{R}^{d\times d}$, we can project metapath-specific embeddings within different metapath-specific tangent space $\mathcal{T}_0\mathbb{D}^{d,c,\phi}$ into same semantic space $\mathbb{R}^{d}$. 

\subsubsection{Metapath aggregation} Given mapped metapath-specific embeddings $g_v^{\phi}$, we aggregate them by using attention mechanisms. First, we calculate the importance of each mapped metapath-specific embedding $\beta_\phi$ as follows:
\begin{align}
    e_\phi &= b^T\cdot\text{tanh}\left(W_3\cdot g_v^\phi+b_3\right),\\
    \beta_\phi &= \frac{\text{exp}\left(e_\phi\right)}{\sum_{\pi\in\Phi}\text{exp}\left(e_\pi\right)}.
\end{align}

In the above equations, $e_\phi$ denotes the importance of each mapped metapath-specific embedding. After calculating $e_\phi$, we normalize these values using the softmax function to obtain their weights. Note that, $W_3 \in \mathbb{R}^{d' \times d}$ denotes a weight matrix for the mapped metapath-specific embeddings, $b_3\in\mathbb{R}^{d'}$ is a bias vector, and $b \in \mathbb{R}^{d'}$ is an attention vector. The embedding vector of node $v$, $z_v\in\mathbb{R}^{d'}$ is calculated as a weighted sum as shown below:
\vspace{-0.5em}
\begin{align}
    z_v&= \sum_{\phi\in\Phi}\left(\beta_\phi\cdot g_v^{\phi}\right).
\end{align}
\subsection{Model Training}
As shown in Equation~(\ref{eq:12}), we employ the non-linear transformation $f(\cdot)$ to map node embedding vectors into a space with the desired output dimension, conducting various downstream tasks. 
\begin{align}
\label{eq:12}
    f\left(z_v\right) &= \sigma\left(W_o\cdot z_v\right),
\end{align}
where $W_o\in\mathbb{R}^{d_o\times d'}$ denotes the weight matrix, $d_o$ denotes the dimension of the output vector, and $\sigma$ is the activation function.
Then, we train \algname{} by minimizing the loss function $\mathcal{L}$.

For node-level tasks, \algname{} is trained by minimizing the cross-entropy loss function $\mathcal{L}_n$ which is defined as below:
\begin{align} 
    \mathcal{L}_n &=-\sum_{v\in V_t}\sum_{c=1}^Cy_v[c]\cdot\text{log}\left(f(z_v)[c]\right),
\end{align}
where $v_t$ is the target node set extracted from the labeled node set, $C$ is the number of classes, $y_v$ is the one-hot encoded label vector for node $v$, and $f(z_v)$ is a vector predicting the label probabilities of node $v$.

For a link-level task, \algname{} is trained by minimizing binary cross-entropy function $\mathcal{L}_l$ which is defined as below:
\begin{align}
\begin{split}
    \mathcal{L}_l  = &-\frac{1}{|\mathcal{S}|}\sum_{\left(u,v,y\right)\in\mathcal{S}}y\cdot \text{log}\left(\sigma\left(z_u^Tz_v\right)\right)\\&+\left(1-y\right)\text{log}\left(1-\sigma\left(z_u^Tz_v\right)\right),
\end{split}
\end{align}
where $\mathcal{S}$ is the set that includes both positive and negative node pairs, $y$ is the ground truth label for a node pair $(u, v)$. $\sigma$ is the sigmoid function, $z_u$ and $z_v$ are the embedding vectors of nodes $u$ and $v$, respectively.

\section{Experiments and Discussion}
In this section, we assess the efficiency of our proposed model \algname{}, through experiments with four real-world heterogeneous graphs. We conduct comparative analyses between \algname{} and several state-of-the-art GNN models. 

\begin{table}[h]
\centering
\caption{Statistics of real-world datasets.}
\vspace{-0.5em}
\label{tab:datastats}
\resizebox{0.9\columnwidth}{!}{%
\begin{tabular}{ccccc} 
\multicolumn{5}{c}{Node Classification and Clustering} \\\hline
Dataset & \# Nodes & \# Links & \# Classes & \# Features \\\hline
IMDB    & \begin{tabular}[c]{@{}c@{}}Movie (M) : 4,661\\ Director (D) : 2,270\\ Actor (A) : 5,841\end{tabular}   & \begin{tabular}[c]{@{}c@{}}M-D : 4,661\\ M-A : 13,983\end{tabular} & 3  & 1,256\\\hline
DBLP    & \begin{tabular}[c]{@{}c@{}}Author (A) : 4,057\\Paper (P) : 14,328\\ Conference (C) : 20\end{tabular}   & \begin{tabular}[c]{@{}c@{}}A-P : 19,645\\ P-C : 14,328\end{tabular} & 4  & 334\\\hline
ACM    & \begin{tabular}[c]{@{}c@{}}Paper (P) : 3,020\\ Author (A) : 5,912\\ Subject (S) : 57\end{tabular}   & \begin{tabular}[c]{@{}c@{}}P-A : 9,936\\ P-S : 3,025\\\end{tabular} & 3 & 1,902\\\hline
\\
\multicolumn{5}{c}{Link Prediction} \\\hline
Dataset & \# Nodes & \# Links & Target & \# Features \\\hline
LastFM    & \begin{tabular}[c]{@{}c@{}}User (U) : 1,892\\ 
Artist (A) : 17,632\\ Tag (T) : 1,088\end{tabular}   & \begin{tabular}[c]{@{}c@{}}U-A : 85,689\\ A-T : 21,553\end{tabular} & User-Artist  & 20,612\\\hline
\end{tabular}%
}
\vspace{-1em}
\end{table}

\subsection{Datasets}
To evaluate the performance of \algname{} on downstream tasks, we use four real-world heterogeneous graph datasets. Table~\ref{tab:datastats} shows the statistics of these datasets. For node classification and clustering tasks, Movie, Author, and Paper types of nodes are labeled.

\begin{table*}[]
\centering
\caption{Experimental results (\%) for the node classification task.}
\vspace{-0.5em}
\label{tab:node_classification}
\resizebox{\textwidth}{!}{%
\begin{tabular}{cccccccccccccccc}
\hline
\multirow{2}{*}{Dataset} &
  \multirow{2}{*}{Metric} &
  \multirow{2}{*}{Train \%} &
  \multicolumn{2}{c}{\lowercase\expandafter{\romannumeral1}} & \multicolumn{1}{c}{\lowercase\expandafter{\romannumeral2}} & \multicolumn{6}{c}{\lowercase\expandafter{\romannumeral3}} & \multicolumn{4}{c}{\lowercase\expandafter{\romannumeral4}}\\ \cmidrule(lr){4-5} \cmidrule(lr){6-6} \cmidrule(lr){7-12} \cmidrule(lr){13-16}
  &
   &
   &
  GCN &
  GAT &
  HGCN &
  HAN &
  MAGNN &
  GTN &
  HGT &
  GraphMSE &
  Simple-HGN &
  McH-HGCN &
  SHAN &
  HHGAT &
  \algname{} \\ \hline
\multirow{8}{*}{IMDB} &
  \multirow{4}{*}{Macro-F1} &
  20\% &
  52.17\footnotesize$\pm$0.35 &
  53.68\footnotesize$\pm$0.26 &
  54.38\footnotesize$\pm$0.48 & 
  56.19\footnotesize$\pm$0.51 &
  59.33\footnotesize$\pm$0.38 &
  58.74\footnotesize$\pm$0.74 &
  56.14\footnotesize$\pm$0.65 &
  57.72\footnotesize$\pm$0.56 &
  59.97\footnotesize$\pm$0.61 &
  58.16\footnotesize$\pm$0.49&
   62.23\footnotesize$\pm$0.76&
   \underline{63.16\footnotesize$\pm$0.39}&
   \bf65.75\footnotesize$\pm$0.81
   \\
 &
   &
  40\% &
  53.20\footnotesize$\pm$0.45 &
  56.33\footnotesize$\pm$0.71 &
  57.05\footnotesize$\pm$0.43&
  56.84\footnotesize$\pm$0.37 &
  60.70\footnotesize$\pm$0.48 &
  59.71\footnotesize$\pm$0.54 &
  57.12\footnotesize$\pm$0.53 &
  62.01\footnotesize$\pm$0.48 &
  61.94\footnotesize$\pm$0.39 &
   60.31\footnotesize$\pm$0.56&
   63.98\footnotesize$\pm$0.68&
   \underline{65.07\footnotesize$\pm$0.63}&
   \bf68.07\footnotesize$\pm$0.54
   \\
 &
   &
  60\% &
  54.35\footnotesize$\pm$0.46 &
  56.93\footnotesize$\pm$0.54 &
   57.86\footnotesize$\pm$0.56&
  58.95\footnotesize$\pm$0.71 &
  60.68\footnotesize$\pm$0.56 &
  61.88\footnotesize$\pm$0.50 &
  61.52\footnotesize$\pm$0.57 &
  65.51\footnotesize$\pm$0.61 &
  \underline{66.73\footnotesize$\pm$0.42}&
   61.93\footnotesize$\pm$0.33&
   66.68\footnotesize$\pm$0.71&
   65.72\footnotesize$\pm$0.56&
   \bf71.42\footnotesize$\pm$0.48
   \\
 &
   &
  80\% &
  54.19\footnotesize$\pm$0.29 &
  57.25\footnotesize$\pm$0.18 &
  57.92\footnotesize$\pm$0.32 &
  58.61\footnotesize$\pm$0.63 &
  61.15\footnotesize$\pm$0.55 &
  62.08\footnotesize$\pm$0.62 &
  63.69\footnotesize$\pm$0.59 &
  67.34\footnotesize$\pm$0.56 &
  67.56\footnotesize$\pm$0.45 &
   62.29\footnotesize$\pm$0.50&
   \underline{68.49\footnotesize$\pm$0.67}&
   67.42\footnotesize$\pm$0.51&
   \bf70.03\footnotesize$\pm$0.60
   \\ \cline{2-16} 
 &
  \multirow{4}{*}{Micro-F1} &
  20\% &
  52.13\footnotesize$\pm$0.38 &
  53.67\footnotesize$\pm$0.31 &
   54.46\footnotesize$\pm$0.42&
  56.71\footnotesize$\pm$0.53 &
  58.30\footnotesize$\pm$0.39 &
  61.97\footnotesize$\pm$0.63 &
  57.97\footnotesize$\pm$0.76 &
  60.58\footnotesize$\pm$0.62 &
  63.76\footnotesize$\pm$0.60 &
   61.28\footnotesize$\pm$0.37&
   64.31\footnotesize$\pm$0.82&
   \underline{65.76\footnotesize$\pm$0.66}&
   \bf69.09\footnotesize$\pm$0.93
   \\
 &
   &
  40\% &
  53.34\footnotesize$\pm$0.41 &
  53.99\footnotesize$\pm$0.65 &
  57.02\footnotesize$\pm$0.46 &
  56.68\footnotesize$\pm$0.70 &
  58.34\footnotesize$\pm$0.58 &
  62.10\footnotesize$\pm$0.53 &
  58.80\footnotesize$\pm$0.65 &
  64.87\footnotesize$\pm$0.63 &
  65.60\footnotesize$\pm$0.46 &
   63.09\footnotesize$\pm$0.12&
   \underline{66.56\footnotesize$\pm$0.73}&
   66.34\footnotesize$\pm$0.70&
   \bf70.95\footnotesize$\pm$0.54
   \\
 &
   &
  60\% &
  54.61\footnotesize$\pm$0.42 &
  56.26\footnotesize$\pm$0.51 &
  58.01\footnotesize$\pm$0.50 &
  58.26\footnotesize$\pm$0.82 &
  60.71\footnotesize$\pm$0.70 &
  63.55\footnotesize$\pm$0.39 &
  62.63\footnotesize$\pm$0.58 &
  68.86\footnotesize$\pm$0.86 &
  69.29\footnotesize$\pm$0.74 &
   64.16\footnotesize$\pm$0.29&
   69.57\footnotesize$\pm$0.76&
   \underline{70.40\footnotesize$\pm$0.51}&
   \bf 73.60\footnotesize$\pm$0.39
   \\
 &
   &
  80\% &
  54.37\footnotesize$\pm$0.33 &
  57.23\footnotesize$\pm$0.29 &
  58.54\footnotesize$\pm$0.93 &
  59.35\footnotesize$\pm$0.65 &
  61.70\footnotesize$\pm$0.39 &
  65.57\footnotesize$\pm$0.91 &
  67.01\footnotesize$\pm$0.47 &
  69.54\footnotesize$\pm$0.59 &
  69.35\footnotesize$\pm$0.66 &
   64.96\footnotesize$\pm$0.42&
  69.42\footnotesize$\pm$0.56 &
   \underline{69.61\footnotesize$\pm$0.89}&
   \bf73.37\footnotesize$\pm$0.59
   \\ \hline
\multirow{8}{*}{DBLP} &
  \multirow{4}{*}{Macro-F1} &
  20\% &
  87.51\footnotesize$\pm$0.15 &
  91.52\footnotesize$\pm$0.34 &
  91.69\footnotesize$\pm$0.38 &
  92.63\footnotesize$\pm$0.46 &
  93.21\footnotesize$\pm$0.64 &
  92.45\footnotesize$\pm$0.37 &
  90.36\footnotesize$\pm$0.62 &
  93.80\footnotesize$\pm$0.39 &
  93.48\footnotesize$\pm$0.56 &
   90.63\footnotesize$\pm$0.72&
  \underline{94.27\footnotesize$\pm$0.16}&
   94.19\footnotesize$\pm$0.08&
  \bf95.44\footnotesize$\pm$0.17
   \\
 &
   &
  40\% &
  88.55\footnotesize$\pm$0.46 &
  91.07\footnotesize$\pm$0.39 &
  91.93\footnotesize$\pm$0.35 &
  92.35\footnotesize$\pm$0.64 &
  93.51\footnotesize$\pm$0.29 &
  92.39\footnotesize$\pm$0.41 &
  91.57\footnotesize$\pm$0.29 &
  94.02\footnotesize$\pm$0.50 &
  93.98\footnotesize$\pm$0.27 &
   91.74\footnotesize$\pm$0.62&
  \underline{94.33\footnotesize$\pm$0.08}&
  94.27\footnotesize$\pm$0.10&
  \bf 95.54\footnotesize$\pm$0.12
   \\
 &
   &
  60\% &
  89.44\footnotesize$\pm$0.27 &
  91.51\footnotesize$\pm$0.46 &
  92.60\footnotesize$\pm$0.89&
  92.86\footnotesize$\pm$0.37 &
  93.59\footnotesize$\pm$0.60 &
  93.77\footnotesize$\pm$0.52 &
  92.32\footnotesize$\pm$0.19 &
  94.30\footnotesize$\pm$0.26 &
  94.01\footnotesize$\pm$0.33 &
   92.26\footnotesize$\pm$0.19&
   94.50\footnotesize$\pm$0.29&
   \underline{94.90\footnotesize$\pm$0.30}&
   \bf 95.67\footnotesize$\pm$0.40
   \\
 &
   &
  80\% &
  89.45\footnotesize$\pm$0.36 &
  91.77\footnotesize$\pm$0.27 &
  92.58\footnotesize$\pm$0.39 &
  92.73\footnotesize$\pm$0.66 &
  94.36\footnotesize$\pm$0.43 &
  94.46\footnotesize$\pm$0.60 &
  93.46\footnotesize$\pm$0.55 &
  94.21\footnotesize$\pm$0.82 &
  94.25\footnotesize$\pm$0.57 &
   93.13\footnotesize$\pm$0.24&
  94.67\footnotesize$\pm$0.12&
  \underline{94.77\footnotesize$\pm$0.19}&
  \bf 95.29\footnotesize$\pm$0.15
   \\ \cline{2-16} 
 &
  \multirow{4}{*}{Micro-F1} &
  20\% &
  88.21\footnotesize$\pm$0.26 &
  91.29\footnotesize$\pm$0.31 &
  92.06\footnotesize$\pm$0.33 &
  92.35\footnotesize$\pm$0.51 &
  93.60\footnotesize$\pm$0.59 &
  93.15\footnotesize$\pm$0.48 &
  91.46\footnotesize$\pm$0.77 &
  94.15\footnotesize$\pm$0.42 &
  94.17\footnotesize$\pm$0.47 &
   92.01\footnotesize$\pm$0.53&
   94.53\footnotesize$\pm$0.17&
   \underline{94.66\footnotesize$\pm$0.07}&
  \bf 95.79\footnotesize$\pm$0.16
   \\
 &
   &
  40\% &
  88.68\footnotesize$\pm$0.52 &
  91.60\footnotesize$\pm$0.50 &
  92.31\footnotesize$\pm$0.40 &
  92.87\footnotesize$\pm$0.39 &
  93.75\footnotesize$\pm$0.44 &
  93.80\footnotesize$\pm$0.56 &
  92.05\footnotesize$\pm$0.48 &
  94.32\footnotesize$\pm$0.81 &
  93.87\footnotesize$\pm$0.42 &
   92.73\footnotesize$\pm$0.51&
   94.60\footnotesize$\pm$0.22&
   \underline{94.72\footnotesize$\pm$0.10}&
  \bf 95.90\footnotesize$\pm$0.11
   \\
 &
   &
  60\% &
  90.01\footnotesize$\pm$0.48 &
  92.09\footnotesize$\pm$0.41 &
  93.16\footnotesize$\pm$0.36 &
  93.42\footnotesize$\pm$0.12 &
  94.20\footnotesize$\pm$0.51 &
  94.22\footnotesize$\pm$0.51 &
  92.72\footnotesize$\pm$0.24 &
  94.38\footnotesize$\pm$0.31 &
  94.71\footnotesize$\pm$0.56 &
   93.50\footnotesize$\pm$0.26&
   94.92\footnotesize$\pm$0.35&
   \underline{95.15\footnotesize$\pm$0.36}&
   \bf 95.98\footnotesize$\pm$0.36
   \\
 &
   &
  80\% &
  90.14\footnotesize$\pm$0.39 &
  92.39\footnotesize$\pm$0.41 &
  93.21\footnotesize$\pm$0.35 &
  93.54\footnotesize$\pm$0.60 &
  94.09\footnotesize$\pm$0.52 &
  94.23\footnotesize$\pm$0.54 &
  92.57\footnotesize$\pm$0.72 &
  94.54\footnotesize$\pm$0.63 &
  94.68\footnotesize$\pm$0.55 &
   93.31\footnotesize$\pm$0.12&
   \underline{95.36\footnotesize$\pm$0.23}&
   95.34\footnotesize$\pm$0.17&
  \bf 95.85\footnotesize$\pm$0.16
   \\ \hline
\multirow{8}{*}{ACM} &
  \multirow{4}{*}{Macro-F1} &
  20\% &
  83.08\footnotesize$\pm$0.37 &
  86.14\footnotesize$\pm$0.49 &
  87.29\footnotesize$\pm$1.06 &
  87.88\footnotesize$\pm$0.42 &
  88.43\footnotesize$\pm$0.51 &
  91.10\footnotesize$\pm$0.39 &
  89.12\footnotesize$\pm$0.46 &
  92.13\footnotesize$\pm$0.27&
  92.25\footnotesize$\pm$0.39 &
   89.86\footnotesize$\pm$0.83&
   \underline{92.56\footnotesize$\pm$0.21}&
   91.34\footnotesize$\pm$0.39&
  \bf 92.73\footnotesize$\pm$0.52
   \\
 &
   &
  40\% &
  87.34\footnotesize$\pm$0.41 &
  87.11\footnotesize$\pm$0.22 &
  89.19\footnotesize$\pm$0.72 &
  90.54\footnotesize$\pm$0.08 &
  90.16\footnotesize$\pm$0.91&
  91.34\footnotesize$\pm$0.44 &
  89.15\footnotesize$\pm$0.49 &
  92.76\footnotesize$\pm$0.37 &
  92.64\footnotesize$\pm$0.61 &
   90.52\footnotesize$\pm$0.69&
   92.88\footnotesize$\pm$0.19&
   \underline{92.92\footnotesize$\pm$0.32}&
  \bf 93.95\footnotesize$\pm$0.51
   \\
 &
   &
  60\% &
  88.80\footnotesize$\pm$0.51 &
  88.92\footnotesize$\pm$0.36 &
  90.01\footnotesize$\pm$0.42 &
  91.22\footnotesize$\pm$0.36 &
  90.73\footnotesize$\pm$0.39 &
  91.34\footnotesize$\pm$0.26 &
  90.57\footnotesize$\pm$0.34 &
  93.39\footnotesize$\pm$0.28 &
  93.06\footnotesize$\pm$0.22 &
   91.03\footnotesize$\pm$0.76&
   94.10\footnotesize$\pm$0.37&
   \underline{94.28\footnotesize$\pm$0.35}&
   \bf 94.83\footnotesize$\pm$0.16
   \\
 &
   &
  80\% &
  88.43\footnotesize$\pm$0.29 &
  88.06\footnotesize$\pm$0.16 &
  90.03\footnotesize$\pm$0.77 &
  91.35\footnotesize$\pm$0.45 &
  92.12\footnotesize$\pm$0.51 &
  91.14\footnotesize$\pm$0.78 &
  93.45\footnotesize$\pm$0.65 &
  93.57\footnotesize$\pm$0.49 &
  93.55\footnotesize$\pm$0.44 &
   91.97\footnotesize$\pm$0.55&
  \bf 94.94\footnotesize$\pm$0.62&
   93.91\footnotesize$\pm$0.31&
   \underline{94.01\footnotesize$\pm$0.30}
   \\ \cline{2-16} 
 &
  \multirow{4}{*}{Micro-F1} &
  20\% &
  87.75\footnotesize$\pm$0.33 &
  87.83\footnotesize$\pm$0.47 &
  88.09\footnotesize$\pm$0.89 &
  91.20\footnotesize$\pm$0.46 &
  91.37\footnotesize$\pm$0.42 &
  91.86\footnotesize$\pm$0.40 &
  89.59\footnotesize$\pm$0.37 &
  92.27\footnotesize$\pm$0.36 &
  91.91\footnotesize$\pm$0.33 &
   90.21\footnotesize$\pm$0.61&
   \underline{92.38\footnotesize$\pm$0.18}&
   92.36\footnotesize$\pm$0.37&
 \bf  92.96\footnotesize$\pm$0.54
   \\
 &
   &
  40\% &
  87.86\footnotesize$\pm$0.42 &
  87.39\footnotesize$\pm$0.41 &
  90.06\footnotesize$\pm$0.73 &
  91.78\footnotesize$\pm$0.28 &
  92.60\footnotesize$\pm$0.48 &
  91.89\footnotesize$\pm$0.46 &
  90.70\footnotesize$\pm$0.43 &
  93.05\footnotesize$\pm$0.36 &
  92.86\footnotesize$\pm$0.84 &
   90.63\footnotesize$\pm$0.52&
   93.37\footnotesize$\pm$0.26&
   \underline{93.46\footnotesize$\pm$0.50}&
   \bf93.91\footnotesize$\pm$0.48
   \\
 &
   &
  60\% &
  88.40\footnotesize$\pm$0.56 &
  87.78\footnotesize$\pm$0.33 &
  90.51\footnotesize$\pm$0.63 &
  92.39\footnotesize$\pm$0.42 &
  92.21\footnotesize$\pm$0.18 &
  92.07\footnotesize$\pm$0.48 &
  91.18\footnotesize$\pm$0.15 &
  93.38\footnotesize$\pm$0.47 &
  93.33\footnotesize$\pm$0.21 &
   91.20\footnotesize$\pm$0.48&
   \underline{94.46\footnotesize$\pm$0.35}&
   94.34\footnotesize$\pm$0.39&
   \bf 94.88\footnotesize$\pm$0.16
   \\
 &
   &
  80\% &
  88.56\footnotesize$\pm$0.33 &
  87.87\footnotesize$\pm$0.51 &
  91.10\footnotesize$\pm$0.44 &
  92.03\footnotesize$\pm$0.16 &
  92.14\footnotesize$\pm$0.48 &
  92.21\footnotesize$\pm$0.66 &
  91.77\footnotesize$\pm$0.57 &
  93.37\footnotesize$\pm$0.36 &
  93.53\footnotesize$\pm$0.42 &
   92.06\footnotesize$\pm$0.66&
  \bf 94.56\footnotesize$\pm$0.11&
   93.72\footnotesize$\pm$0.32&
   \underline{94.05\footnotesize$\pm$0.31}
   
   \\ \hline
\end{tabular}%
}
\end{table*}

\begin{table*}[]
\centering
\caption{Experimental results (\%) for the node clustering task.}
\vspace{-0.5em}
\label{tab:node_clustering}
\resizebox{\textwidth}{!}{%
\begin{tabular}{ccccccccccccccc}
\hline
\multirow{2}{*}{Dataset} &
  \multirow{2}{*}{Metric} &
  \multicolumn{2}{c}{\lowercase\expandafter{\romannumeral1}} & \multicolumn{1}{c}{\lowercase\expandafter{\romannumeral2}} & \multicolumn{6}{c}{\lowercase\expandafter{\romannumeral3}} & \multicolumn{4}{c}{\lowercase\expandafter{\romannumeral4}}\\ \cmidrule(lr){3-4} \cmidrule(lr){5-5} \cmidrule(lr){6-11} \cmidrule(lr){12-15}
  &
   &
  GCN &
  GAT &
  HGCN &
  HAN &
  MAGNN &
  GTN &
  HGT &
  GraphMSE &
  Simple-HGN &
  McH-HGCN &
  SHAN &
  HHGAT &
  \algname{} \\ \hline
\multirow{2}{*}{IMDB} &
  NMI &
  7.84\footnotesize$\pm$0.24 &
  8.06\footnotesize$\pm$0.18 &
  10.29\footnotesize$\pm$0.76 &
  11.21\footnotesize$\pm$1.09 &
  15.66\footnotesize$\pm$0.73 &
  15.01\footnotesize$\pm$0.11 &
  14.55\footnotesize$\pm$0.32 &
  15.70\footnotesize$\pm$1.25 &
  17.58\footnotesize$\pm$0.82 &
  14.32\footnotesize$\pm$0.46 &
  20.60\footnotesize$\pm$0.92 &
  \underline{20.75\footnotesize$\pm$0.36}&
  \bf24.06\footnotesize$\pm$0.51\\
 &
  ARI &
  8.12\footnotesize$\pm$0.40 &
  8.86\footnotesize$\pm$0.09 &
  11.10\footnotesize$\pm$0.88 &
  11.49\footnotesize$\pm$0.11 &
  16.72\footnotesize$\pm$0.21 &
  15.96\footnotesize$\pm$0.63 &
  16.59\footnotesize$\pm$0.36 &
  16.38\footnotesize$\pm$0.74 &
  19.51\footnotesize$\pm$1.06 &
  16.91\footnotesize$\pm$0.37 &
  22.56\footnotesize$\pm$0.22 &
  \underline{22.80\footnotesize$\pm$0.68} &
  \bf 26.33\footnotesize$\pm$0.46\\ \hline
\multirow{2}{*}{DBLP} &
  NMI &
  75.37\footnotesize$\pm$0.25 &
  75.46\footnotesize$\pm$0.44 &
  76.48\footnotesize$\pm$0.87 &
  77.03\footnotesize$\pm$0.16 &
  80.11\footnotesize$\pm$0.30 &
  81.39\footnotesize$\pm$0.73 &
  79.02\footnotesize$\pm$0.39 &
  37.22\footnotesize$\pm$0.65 &
  82.38\footnotesize$\pm$0.07 &
  78.90\footnotesize$\pm$0.31 &
  82.39\footnotesize$\pm$0.42 &
  \underline{83.14\footnotesize$\pm$0.19} &
  \bf84.38\footnotesize$\pm$0.59\\
 &
  ARI &
  77.14\footnotesize$\pm$0.21 &
  77.99\footnotesize$\pm$0.72 &
  79.36\footnotesize$\pm$0.95 &
  82.53\footnotesize$\pm$0.42 &
  85.61\footnotesize$\pm$0.38 &
  84.12\footnotesize$\pm$0.83 &
  80.28\footnotesize$\pm$0.20 &
  34.21\footnotesize$\pm$0.65 &
  85.71\footnotesize$\pm$0.33 &
  81.22\footnotesize$\pm$0.56 &
  \underline{86.13\footnotesize$\pm$0.33} &
  85.91\footnotesize$\pm$0.49 &
  \bf88.27\footnotesize$\pm$0.63\\ \hline
\multirow{2}{*}{ACM} &
  NMI &
  51.73\footnotesize$\pm$0.21 &
  58.06\footnotesize$\pm$0.46 &
  60.19\footnotesize$\pm$0.69 &
  61.24\footnotesize$\pm$0.12 &
  64.73\footnotesize$\pm$0.47 &
  65.06\footnotesize$\pm$0.35 &
  67.88\footnotesize$\pm$0.20 &
  66.65\footnotesize$\pm$0.44 &
  69.91\footnotesize$\pm$0.68 &
  66.76\footnotesize$\pm$0.38 &
  \underline{72.90\footnotesize$\pm$0.93}&
  72.49\footnotesize$\pm$0.44 &
  \bf73.33\footnotesize$\pm$0.79\\
 &
  ARI &
  53.42\footnotesize$\pm$0.48 &
  59.61\footnotesize$\pm$0.42 &
  62.06\footnotesize$\pm$0.70 &
  64.11\footnotesize$\pm$0.26 &
  66.84\footnotesize$\pm$0.25 &
  65.80\footnotesize$\pm$0.49 &
  72.56\footnotesize$\pm$0.13 &
  73.89\footnotesize$\pm$0.33 &
  72.07\footnotesize$\pm$0.51 &
  71.84\footnotesize$\pm$0.48 &
  77.73\footnotesize$\pm$0.44 &
  \underline{77.92\footnotesize$\pm$0.80} &
  \bf78.28\footnotesize$\pm$ 1.07 \\ \hline
\end{tabular}%
}
\end{table*}

\begin{table*}[]
\centering
\caption{Experimental results (\%) for the link prediction task.}
\vspace{-0.5em}
\label{tab:link_prediction}
\resizebox{0.8\textwidth}{!}{%
\begin{tabular}{cccccccccccc}
\hline
\multirow{2}{*}{Dataset} &
  \multirow{2}{*}{Metric} &
  \multicolumn{2}{c}{\lowercase\expandafter{\romannumeral1}} & \multicolumn{1}{c}{\lowercase\expandafter{\romannumeral2}} & \multicolumn{5}{c}{\lowercase\expandafter{\romannumeral3}} & \multicolumn{2}{c}{\lowercase\expandafter{\romannumeral4}}\\ \cmidrule(lr){3-4} \cmidrule(lr){5-5} \cmidrule(lr){6-10} \cmidrule(lr){11-12}
  &
   &
  GCN &
  GAT &
  HGCN &
  HAN &
  MAGNN &
  HetSANN &
  HGT &
  Simple-HGN &
  HHGAT &
  \algname{} \\ \hline
\multirow{2}{*}{LastFM} &
  ROC-AUC &
  43.68\footnotesize$\pm$0.30 &
  44.52\footnotesize$\pm$0.22 &
  46.71\footnotesize$\pm$0.78 &
  48.32\footnotesize$\pm$0.28 &
  49.37\footnotesize$\pm$0.59 &
  50.28\footnotesize$\pm$0.45 &
  47.78\footnotesize$\pm$0.23 &
  53.85\footnotesize$\pm$0.47 &
  \underline{54.37\footnotesize$\pm$0.51}&
  \bf 55.77\footnotesize$\pm$0.62 \cr
 &
  F1-Score &
   56.15\footnotesize$\pm$0.16 &
  56.84\footnotesize$\pm$0.07 &
  57.23\footnotesize$\pm$0.66 &
  57.11\footnotesize$\pm$0.49 &
  58.37\footnotesize$\pm$0.32 &
  60.61\footnotesize$\pm$0.54 &
  61.16\footnotesize$\pm$0.57 &
  \underline{63.02\footnotesize$\pm$0.35} &
  62.85\footnotesize$\pm$0.48 &
  \bf 63.39\footnotesize$\pm$0.76 \cr
  \hline
\end{tabular}%
}
\vspace{-1em}
\end{table*}

\subsection{Baselines}
We compare \algname{} with several state-of-the-art graph neural networks categorized into four groups- \textbf{\lowercase\expandafter{\romannumeral1}) Euclidean homogeneous GNNs}: GCN~\cite{kips2017iclr} and GAT~\cite{velickovic2018iclr}, \textbf{\lowercase\expandafter{\romannumeral2}) Hyperbolic homogeneous GNNs}: HGCN~\cite{chami2019hyperbolic}, \textbf{\lowercase\expandafter{\romannumeral3}) Euclidean heterogeneous GNNs}: HAN~\cite{wang2019heterogeneous}, MAGNN~\cite{fu2020magnn}, GTN~\cite{yun2019graph}, HetSANN~\cite{hong2020attention}, HGT~\cite{hu2020heterogeneous}, GraphMSE~\cite{li2021graphmse}, and Simple-HGN~\cite{lv2021we}, and \textbf{\lowercase\expandafter{\romannumeral4}) Hyperbolic heterogeneous GNNs}: McH-HGCN~\cite{liu2023mch}, SHAN~\cite{li2023multi}, and HHGAT~\cite{park2024hyperbolic}. For homogeneous GNNs, features are processed to be homogeneous for pair comparison with heterogeneous GNNs. 

\subsection{Node Classification and Clustering } Node classification was performed by applying support vector machines on embedding vectors of labeled nodes. Macro-F1 and Micro-F1 were used as the evaluation metrics for classification accuracy. The ratio of training data was varied within the range of 20\% to 80\%. For node clustering, the $k$-means clustering algorithm was applied to embedding vectors of labeled nodes. Normalized Mutual Information (NMI) and Adjusted Rand Index (ARI) were used as the evaluation metrics for clustering accuracy.

As shown in Table~\ref{tab:node_classification} and~\ref{tab:node_clustering}, \algname{} achieved better performance than other baselines in most cases. The results from \algname{} and HHGAT indicate the effectiveness of using multi-hyperbolic space to learn metapath instances. Specifically, because HHGAT employs a single hyperbolic space to learn metapath instances, it cannot effectively capture the complex structures following various power-law distributions of metapath instances in a single hyperbolic space. In contrast, \algname{} effectively captures various complex structures by using multi-hyperbolic space to learn metapath instances. In each metapath-specific hyperbolic space corresponding to a distinct metapath, a learned negative curvature effectively represents the distribution of node degrees for metapath instances following that metapath. Furthermore, through intra-hyperbolic space attention and inter-hyperbolic space attention, \algname{} effectively captures important complex structures and semantic information within a heterogeneous graph, respectively. Moreover, a comparison of \algname{} and McH-HGCN demonstrates that \algname{} effectively captures a broader range of semantic information and structural properties by extensively sampling the surrounding structure of the target node, in contrast to McH-HGCN.

On the one hand, a comparison of Euclidean homogeneous GNNs and hyperbolic homogeneous GNNs demonstrates the effectiveness of hyperbolic space in representing complex structures within heterogeneous graphs. However, from a comparison of hyperbolic homogeneous GNNs and Euclidean heterogeneous GNNs, we observe that while hyperbolic homogeneous GNNs can learn complex structures, they struggle to capture the heterogeneity within heterogeneous graphs, thus failing to learn semantic information effectively. In contrast, Euclidean heterogeneous GNNs excel at learning semantic information within such graphs. At last, comparing Euclidean heterogeneous GNNs and hyperbolic heterogeneous GNNs, we can conclude that hyperbolic heterogeneous GNNs are more effective than Euclidean heterogeneous GNNs because hyperbolic heterogeneous GNNs can simultaneously learn the complex structures and heterogeneity of heterogeneous graphs.

\begin{table*}[]
\centering
\caption{Results of the ablation study.}
\vspace{-0.5em}
\label{tab:ablation}
\resizebox{\textwidth}{!}{%
\begin{tabular}{ccccccccccccc}
\hline
Dataset & \multicolumn{4}{c}{IMDB} & \multicolumn{4}{c}{DBLP} & \multicolumn{4}{c}{ACM} \\ \cmidrule(lr){2-5} \cmidrule(lr){6-9} \cmidrule(lr){10-13}
Metric & Macro-F1 & Micro-F1 & NMI & ARI & Macro-F1 & Micro-F1 & NMI & ARI & Macro-F1 & Micro-F1 & NMI & ARI \\ \hline
\algname{} & \bf71.42\footnotesize$\pm$0.48 & \bf73.60\footnotesize$\pm$0.39  & \bf24.06\footnotesize$\pm$0.51 & \bf26.33\footnotesize$\pm$0.46 & \bf95.67\footnotesize$\pm$0.40 & \bf95.98\footnotesize$\pm$0.36 & \bf84.38\footnotesize$\pm$0.59 & \bf88.27\footnotesize$\pm$0.63 & \bf94.83\footnotesize$\pm$0.16 & \bf94.88\footnotesize$\pm$0.16 & \bf73.33\footnotesize$\pm$0.79 & \bf78.28\footnotesize$\pm$1.07    \\
\algname{}$_{\text{CONCAT}}$ & 68.03\footnotesize$\pm$0.44 &  70.75\footnotesize$\pm$0.44 & 22.24\footnotesize$\pm$0.55 & 24.91\footnotesize$\pm$0.53 & 94.23\footnotesize$\pm$0.50 & 94.54\footnotesize$\pm$0.47 & 82.06\footnotesize$\pm$0.79 & 84.65\footnotesize$\pm$0.88 & 93.33\footnotesize$\pm$0.07 & 93.91\footnotesize$\pm$0.26 & 72.73\footnotesize$\pm$0.44 & 75.82\footnotesize$\pm$0.27 \\
\algname{}$_{\text{EUCLID}}$ & 64.72\footnotesize$\pm$0.10& 67.17\footnotesize$\pm$0.12 & 16.72\footnotesize$\pm$0.18 & 14.65\footnotesize$\pm$0.44 & 93.09\footnotesize$\pm$0.35 & 93.51\footnotesize$\pm$0.33  & 79.32\footnotesize$\pm$0.94 & 83.84\footnotesize$\pm$1.09 & 90.02\footnotesize$\pm$1.15
&  90.09\footnotesize$\pm$1.18  &  70.24\footnotesize$\pm$1.93& 73.18\footnotesize$\pm$2.26  \\
\algname{}$_{\text{SINGLE}}$ & 66.06\footnotesize$\pm$0.56 & 68.77\footnotesize$\pm$0.61  & 21.65\footnotesize$\pm$0.59 & 25.91\footnotesize$\pm$0.80 & 93.76\footnotesize$\pm$0.42 & 93.94\footnotesize$\pm$0.33 & 80.95\footnotesize$\pm$0.32 & 86.06\footnotesize$\pm$0.56 & 92.39\footnotesize$\pm$0.29 & 92.67\footnotesize$\pm$0.26 & 73.16\footnotesize$\pm$0.07 & 77.93\footnotesize$\pm$0.38  \\ \hline
\end{tabular}%
}
\vspace{-1em}
\end{table*}

\subsection{Link Prediction}
We also conducted a link prediction task on the LastFM dataset. To predict the probabilities of relations between user-type nodes and artist-type nodes, we used a dot product operation applied to the embeddings of the two types of nodes. The area under the ROC curve (ROC-AUC) and F1-score were used as the evaluation metrics for prediction accuracy. We considered all connected user-artist pairs as positive samples, while unconnected user-artist pairs were considered as negative samples. For model training, an equal number of positive and negative samples were used.

As shown in Table~\ref{tab:link_prediction}, \algname{} outperforms the other baselines. Compared \algname{} with HHGAT, in link prediction, when predicting the connection between two nodes of different types, the metapath defined around each type of node changes completely, and the distribution of metapath instances also differs. Consequently, while \algname{} can flexibly learn from various metapath instance distributions, HHGAT is unable to do so, making MSGAT superior to HHGAT in link prediction.

\subsection{Ablation Study} We compose three variants of \algname{} to validate the effectiveness of each component of \algname{}. \algname{}$_{\text{CONCAT}}$ concatenates node features within metapath instances, instead of using hyperbolic mean-linear encoder, \algname{}$_{\text{EUCLID}}$  uses Euclidean space for embedding space instead of hyperbolic space, and \algname{}$_{\text{SINGLE}}$ uses only one hyperbolic space to learn metapath instance embeddings. Note that, the training percentage for node classification is set to 60\%. 

We report the results of the ablation study in Table~\ref{tab:ablation}. Comparing \algname{} with \algname{}$_{\text{CONCAT}}$, we observe that transforming node features to the metapath-specific hyperbolic space is more effective than simply concatenating node features within the metapath instance. Next, a comparison of \algname{}$_{\text{EUCLID}}$ and \algname{}$_{\text{SINGLE}}$ demonstrates that using hyperbolic space as the embedding space is more effective for learning heterogeneous graphs than using Euclidean space. Moreover, comparing \algname{} with \algname{}$_{\text{SINGLE}}$, we can see that the use of multi-hyperbolic space leads to significant performance improvements. This is due to the ability of \algname{} to effectively learn the complex structures of diverse distributions within heterogeneous graphs.

\section{Conclusion}

In this paper, we propose the Multi-hyperbolic Space-based heterogeneous Graph Attention Network (\algname{}). Instead of the Euclidean space, \algname{} uses multiple hyperbolic spaces to capture various power-law structures effectively, and finally \algname{} aggregates metapath-specific embeddings to obtain more enhanced node representations.

We conduct comprehensive experiments to evaluate the effectiveness of \algname{} with widely used real-world heterogeneous graph datasets. 
The experimental results demonstrate that \algname{} outperforms the other state-of-the-art baselines. Additionally, it has been shown that using multiple hyperbolic spaces for learning various power-law distributions is effective. 

For the future work, we plan to develop methods to enhance the interpretability of the hyperbolic spaces learned for each metapath in heterogeneous graphs.

\vspace{-0.25em}
\section*{Acknowledgement}
This work was supported by the National Research Foundation of Korea (NRF) grant funded by the Korea government (MSIT) (No.RS-2023-00214065) and by the Institute of Information \& Communications Technology Planning \& Evaluation (IITP) grant funded by the Korea government (MSIT) (No.RS-2022 00155857, Artificial Intelligence Convergence Innovation Human Resources Development (Chungnam National University)).
\vspace{-0.25em}
\bibliographystyle{IEEEtran}
\bibliography{citations}

\end{document}